\documentclass[NewProceedings]{ascelike-new}

\NameTag{Eugene Denteh \today}

\usepackage{pifont} 
\usepackage[utf8]{inputenc}
\usepackage[T1]{fontenc}
\usepackage{lmodern}
\usepackage{graphicx}
\usepackage[style=base,figurename=Fig.,labelfont=bf,labelsep=period]{caption}
\usepackage{subcaption}
\usepackage{amsmath}
\usepackage{newtxtext,newtxmath}
\usepackage[colorlinks=true,citecolor=red,linkcolor=black]{hyperref}
\usepackage{booktabs} 
\usepackage{multirow}

\hypersetup{
    colorlinks=true,
    linkcolor=black,
    citecolor=black,
    urlcolor=black
}
\begin{document}

\title{Hybrid Congestion Classification Framework Using Flow-Guided Attention and Empirical Mode Decomposition}

\author[1]{Eugene Kofi Okrah Denteh}
\author[2]{Blessing Agyei Kyem}
\author[3]{Joshua Kofi Asamoah}
\author[4]{Armstrong Aboah}

\affil[1]{Civil, Construction and Environmental Engineering Department, North Dakota State University, North Dakota, United States of America, Email: eugene.denteh@ndsu.edu (Corresponding Author)}
\affil[2]{Civil, Construction and Environmental Engineering Department, North Dakota State University, North Dakota, United States of America, blessing, Email: blessing.agyeikyem@ndsu.edu}
\affil[3]{Civil, Construction and Environmental Engineering Department, North Dakota State University, North Dakota, United States of America, Email: joshua.asamoah@ndsu.edu}
\affil[4]{Civil, Construction and Environmental Engineering Department, North Dakota State University, North Dakota, United States of America, Email: armstrong.aboah@ndsu.edu}

\maketitle

\begin{abstract}
Accurate traffic congestion classification requires models that jointly capture roadway scene context and non-stationary traffic motion, yet most prior work treats these requirements in isolation. Vision-based methods often depend on appearance cues with standard temporal pooling, which can bias predictions toward static infrastructure, whereas signal-based approaches characterize temporal dynamics but lack the spatial context needed for scene-level localization. These complementary limitations motivate a unified framework that links motion evidence to spatial feature selection while preserving data-adaptive temporal characterization. This study therefore proposes FLO-EMD, a hybrid approach that couples motion-guided attention with empirical, data-driven temporal decomposition. Dense optical flow guides channel and spatial attention so that RGB features are refined toward motion-relevant regions. In parallel, aggregated flow statistics form compact motion traces that are decomposed using Empirical Mode Decomposition (EMD) to extract intrinsic temporal components. The resulting EMD embedding is fused with learned spatiotemporal representations to classify light, medium, and heavy congestion. Experiments on 1,050 five-second clips from four surveillance networks show that FLO-EMD achieves 97.5\% overall test accuracy (weighted F1 = 0.9742), outperforming established baselines and remaining robust across diverse environmental conditions; ablation and sensitivity analyses further quantify the contributions of EMD, the number of intrinsic mode functions, and the selected motion descriptors.
\end{abstract}

\section{Practical Applications}
This research introduces Flow-Guided Empirical Mode Decomposition (FLO-EMD), a traffic congestion classification system that uses existing surveillance camera footage to automatically categorize traffic conditions as light, medium, or heavy, achieving 97.5\% accuracy. By leveraging deployed closed-circuit television (CCTV) infrastructure, transportation agencies can expand monitoring coverage without installing additional roadside sensors. Because the method is explicitly motion-driven, it supports reliable operation under adverse weather and low-light conditions. The flow-guided attention mechanism also limits sensitivity to static scene artifacts, improving robustness in complex scenes. As a result, Flow-Guided Empirical Mode Decomposition is well suited for routine corridor monitoring and real-time operational decision support within intelligent transportation systems.

\section{Introduction}
Traffic congestion has become one of the most pressing challenges facing modern urban transportation systems worldwide. This occurs when vehicle demand exceeds the available road capacity, resulting in reduced speeds, longer travel times, and increased vehicle density along transportation corridors. This imposes significant economic and operational losses through delayed commutes, increased fuel consumption, and reduced productivity \cite{schrank20072007,reedinrix,kyem2025advancing,agyei2026pavecap}. To manage congestion, numerous studies have developed various approaches, broadly categorized into sensor-based and vision-based methodologies \cite{thabit2024survey}. Sensor-based systems employ technologies such as inductive loop detectors, radar sensors, and GPS tracking devices to collect quantitative traffic data including vehicle speeds, flow rates, and occupancy measurements at specific locations \cite{klein2006traffic,adm2016traffic}. In this setting, traffic data is commonly analyzed as time-varying signals, where signal decomposition techniques such as Empirical Mode Decomposition (EMD) \cite{huang1998empirical} can isolate intrinsic temporal components and capture non-stationary variations in flow dynamics without requiring fixed frequency assumptions. While these systems are effective in highway monitoring applications, they face significant limitations in spatial coverage because they monitor discrete locations rather than providing comprehensive area-wide awareness \cite{SOUZA2022103292,kyem2026self,denteh2025integrating,kyem2025task}. 

To address coverage limitations, many studies have turned to vision-based approaches that leverage surveillance-camera footage to observe traffic conditions across broader road segments. Numerous studies have employed deep learning models such as convolutional neural networks for extracting spatial scene cues, recurrent architectures for temporal sequence modeling \cite{lin2024mobilenetv2,wang2025traffic,kyem2025pavesync,denteh2025demographics}, and attention-based networks such as the transformer architecture \cite{khalladi2024efficient,kyem2025big} for learning salient spatiotemporal dependencies, enabling congestion classification directly from traffic video and providing richer spatial context than point-based sensors. However, despite their strong spatial coverage, vision-based models often rely on appearance-driven attention and generic temporal aggregation, which can reduce sensitivity to non-stationary motion patterns \cite{li2023mitigating,zhou2025seeing,dentehhybrid} such as stop-and-go oscillations and gradual queue formation. These designs can also encourage attention to drift toward static high-contrast infrastructure rather than motion-relevant regions that directly reflect traffic state. Taken together, the limitations of sensor-based and vision-based approaches indicate that neither paradigm alone is sufficient for reliable congestion classification in real-world deployments. The key challenge is that visually similar frames can correspond to different congestion states when motion patterns differ, so for a model to effectively classify congestion, it should be able to represent both appearance and motion over time. Sensor measurements support temporally adaptive analysis but offer limited spatial context, whereas vision-based models provide area-wide scene coverage but often lack temporally adaptive motion characterization. This motivates a unified framework that integrates both temporal signal structure and spatial scene information to capture how congestion evolves over time and where it manifests within the roadway scene.

Optical flow \cite{xu2022gmflow} offers a practical bridge between appearance cues and motion dynamics because it estimates dense, pixel-level displacement between consecutive frames and therefore directly represents how traffic moves through the scene. It has been widely used to capture motion structure in traffic videos \cite{yin2021deep}, such as local velocity patterns and evolving movement fields, while reducing reliance on static visual appearance. Recent deep optical-flow models further improve robustness and accuracy under real-world conditions, making optical flow a strong front-end motion representation for congestion understanding tasks \cite{shi2024two}.

Therefore, we propose FLO-EMD, a hybrid congestion classification framework that integrates motion signal analysis with spatiotemporal video understanding. The framework treats dense optical flow as the primary representation of traffic motion and uses it in two complementary roles. First, optical flow guides spatial and channel attention so that the model emphasizes motion-relevant roadway regions rather than static infrastructure. Second, statistical summaries of the optical flow field are organized as motion traces over time and decomposed using Empirical Mode Decomposition (EMD) to capture intrinsic, non-stationary temporal components that are difficult to represent with fixed-window temporal modeling. These EMD-derived components complement learned video features, improving robustness under changing environmental conditions while supporting interpretability through motion-aligned attention visualizations. The primary contributions of this work are as follows:
\begin{enumerate}
    \item We present an application of empirical mode decomposition to optical–flow time series, which adaptively decomposes non-stationary traffic motion into intrinsic mode functions. This enables the characterization of multi-scale temporal dynamics that are critical for distinguishing congestion levels, providing a more robust alternative to conventional fixed-window temporal analysis.

    \item We also present a hybrid framework that seamlessly combines classical signal processing (EMD) with modern deep learning techniques (dual encoder architecture and attention mechanisms). This integration bridges spatial video analysis and temporal pattern recognition, delivering a unified system that outperforms existing methods in capturing traffic congestion dynamics.
\end{enumerate}
The remainder of this paper is organized as follows. We review related work, present the proposed FLO-EMD framework, describe the experimental setup, and report results and scenario-based analyses, followed by conclusions and future directions.

\section{Related Work}

In recent years, numerous studies have focused on two complementary streams in traffic congestion research. Sensor-based studies treat congestion as a time-varying process observed through speed, flow, or occupancy measurements and focus on extracting informative temporal structure from non-stationary signals. Conversely, vision-based studies infer congestion from camera streams using spatial and motion cues. This section reviews representative advances in both streams, emphasizing signal-processing and time-series methods for congestion dynamics as well as vision-based spatiotemporal learning approaches for surveillance video analysis.

\subsection{Sensor-Based Congestion Modelling and Data-Adaptive Signal Decomposition}

Sensor-based congestion monitoring relies on loop detectors, radar, and probe trajectories to generate time series describing traffic states at instrumented locations \cite{zhang2025review}. Traditional analyses often apply threshold-based indicators and parametric time-series models; however, congestion evolves under incidents, demand fluctuations, and control actions, producing signals that are non-linear and non-stationary \cite{tak2026early,he2023autonomous,du2025traffic}. Under these conditions, fixed-basis transforms provide an imperfect match for temporal characterization because their assumptions and parameterizations can conflict with time-varying congestion dynamics \cite{tak2026early,he2023autonomous}. Fourier analysis assumes stationarity and offers weak localization for transient congestion events \cite{app152312698}. Wavelet methods improve time--frequency localization, but their performance depends on the selected mother wavelet and scales, which may not align with evolving traffic oscillations \cite{sahoo2024optimal}. As a result, when analyzing speed, flow, or occupancy time series from loop detectors, radar, or probes, fixed-basis transforms can under-represent short-lived breakdown and recovery patterns and can be sensitive to user-selected bases and scales as congestion regimes shift over time.

Empirical Mode Decomposition (EMD) addresses these limitations through a data-adaptive decomposition into intrinsic mode functions without predefined basis functions \cite{huang1998empirical}. Noise-assisted variants, including EEMD \cite{wu2009ensemble} and CEEMDAN \cite{torres2011complete}, further improve robustness by reducing mode mixing and stabilising the decomposition. In transportation applications, these EMD variants are often used as front-end decomposers to isolate multiscale components and strengthen downstream forecasting or detection models, particularly when paired with modern learning architectures \cite{hu2023short,bing2024short,tian2025short}. The key point for congestion analysis is that EMD-family methods improve temporal characterization of non-stationary dynamics, but they are typically applied to 1D point or link measurements. Nonetheless, sensor-based methods remain spatially constrained as point or link measurements cannot directly localize where congestion forms within the roadway scene or represent lane-level heterogeneity across an entire camera field of view \cite{logman2002integration}. For congestion classification and operations, this limits interpretability because the output cannot indicate which parts of the observed roadway segment are congested or how congestion differs by lane within the monitored scene.

\subsection{Vision-Based Congestion Classification, Optical Flow, and Attention Limitations}
Vision-based congestion classification addresses the spatial coverage limitations of point sensors by using surveillance video to observe traffic conditions over extended road segments \cite{thabit2024survey}. Early systems typically relied on hand-crafted visual proxies, such as vehicle counts, density/occupancy surrogates, and queue-related measures derived from predefined regions of interest \cite{kumar2021applications}. As data availability and compute improved, research shifted toward learned spatiotemporal representations, including CNN-based pipelines with attention modules and, more recently, transformer-style video architectures that learn global context through token interactions \cite{lin2024mobilenetv2,khalladi2024efficient}.

A central requirement in this setting is to represent motion in a way that reflects congestion dynamics rather than static appearance. Optical flow is widely used for this purpose because it provides dense, pixel-level displacement fields that track how traffic moves through the scene. By encoding speed variation, directional coherence, and stop-and-go oscillations more directly than RGB cues alone, optical flow helps disambiguate visually similar scenes that exhibit different motion regimes \cite{bouraffa2024comparing,shi2024two}. Recent vision-based congestion studies increasingly employ attention mechanisms and transformer architectures to strengthen spatiotemporal modelling \cite{bertasius2021space,arnab2021vivit,liu2022video}. Attention modules reweight features to emphasise salient regions, while transformers represent video as a sequence of tokens and use self-attention to capture long-range dependencies across space and time. These models enable modelling of broader spatial context and longer temporal dependencies than single-frame or lane-local cues \cite{bertasius2021space}.

However, two limitations persist for congestion classification. First, attention learned primarily from appearance-dominant tokens is not inherently constrained to motion-relevant evidence, so saliency can drift toward static, high-contrast infrastructure and imaging artefacts such as lane markings, barriers, shadows, and glare. This creates spatial misalignment with traffic dynamics and reduces operator-facing interpretability \cite{li2023mitigating}. Second, transformer temporal modelling typically performs generic feature aggregation rather than explicitly decomposing non-stationary motion, which is central to congestion formation and dissipation. As a result, short-lived stop-and-go oscillations can be smoothed over, and gradual regime transitions can be under-emphasised, making it difficult to consistently separate transient fluctuations from sustained congestion states under varying environmental conditions.

\subsection{The Need for Hybrid Signal-Vision Frameworks}
The literature points to complementary strengths and persistent limitations. EMD-based congestion studies generally operate on 1D traffic sensor signals such as speed, flow, and occupancy, using EMD to decompose the time series before a separate forecasting or classification stage, and therefore providing limited access to scene-level spatial context \cite{hu2023short,bing2024short,tian2025short}. In contrast, vision-based approaches offer area-wide spatial coverage and can leverage optical-flow motion evidence, but appearance-dominant attention and generic temporal modelling are not consistently aligned with non-stationary congestion processes and transient stop-and-go oscillations \cite{li2023mitigating}. Together, these findings indicate an open gap for a unified representation that simultaneously captures where congestion forms within the roadway scene and how its motion dynamics evolve under non-stationary conditions.

Motivated by this gap, we propose a unified signal--vision hybrid that couples data-adaptive temporal decomposition with motion-conditioned spatial representation learning. The framework first computes dense optical flow from traffic video and aggregates it into motion-trace time series that reflect scene-level congestion dynamics. It then applies Empirical Mode Decomposition to these video-derived motion traces to represent non-stationary temporal structure in a fully data-adaptive manner. Finally, the resulting decomposition embedding is fused with a flow-guided spatial attention module and a spatiotemporal video encoder within a single architecture. This formulation departs from prior EMD-based congestion studies in two respects, namely the use of video-derived motion traces rather than detector signals and the direct integration of decomposition features with motion-guided attention and spatiotemporal encoding rather than treating EMD as a standalone preprocessing step. Consequently, the proposed hybrid framework preserves spatial context for localization and interpretability while explicitly modelling non-stationary congestion evolution through adaptive decomposition of motion dynamics.

\section{Methodology}
\label{sec:methodology}

\subsection{Framework Overview}
\label{sec:problem_formulation}

Vision-based traffic congestion classification  presents a multimodal classification challenge that requires the integration of spatial features with temporal motion dynamics to accurately distinguish between different traffic states. Specifically, given a traffic clip $\mathbf{V}=\{I_1,\ldots,I_T\}$ with frames $I_t\in\mathbb{R}^{H\times W\times C}$, we learn a mapping $F:\mathbb{R}^{T\times H\times W\times C}\rightarrow\{1,\ldots,K\}$ that assigns the clip to one of $K$ congestion classes. Table~\ref{tab:notation} summarizes the notation used throughout this section. Even when two clips appear visually similar in terms of vehicle density and scene layout, their congestion states can differ due to underlying motion dynamics. Therefore, effective congestion classification requires representations that jointly capture spatial appearance cues and their temporal evolution in traffic motion.

\begin{figure*}[ht]
\centering
\includegraphics[width=1.1\textwidth]{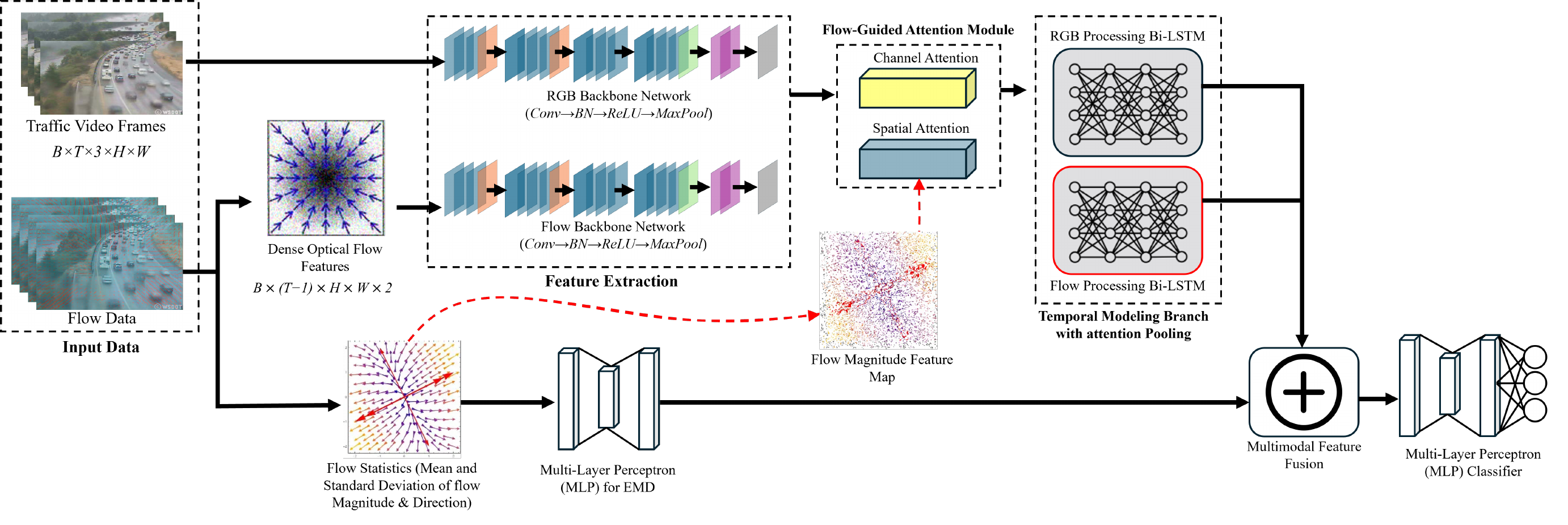}
\caption{Overview of the FLO-EMD framework. The architecture processes traffic video sequences through parallel RGB and optical flow backbones, employs flow-guided attention mechanisms for spatial feature enhancement, integrates EMD-based temporal analysis of motion statistics, and fuses multimodal features through bidirectional LSTM encoding for final classification.}
\label{fig:methodology_overview}
\end{figure*}

Figure~\ref{fig:methodology_overview} presents the FLO-EMD framework. FLO-EMD processes RGB frames and dense optical flow in parallel, uses optical flow to guide spatial and channel attention of RGB features through an enhanced CBAM block, and models temporal dependencies with a bidirectional LSTM module. In addition to learned video representations, FLO-EMD extracts compact frame-level motion statistics from the optical flow field and applies Empirical Mode Decomposition to obtain temporally adaptive components that complement the spatiotemporal features. The final prediction is produced from the fused representation that combines the learned visual embedding and the EMD-derived motion embedding.

The rest of this section details the implementation of each framework component, starting with \textit{Optical Flow Extraction and Motion Characterization}, which describes dense optical flow estimation and the construction of compact motion descriptors used for temporal analysis. Next, \textit{Dual-Stream Backbone Architecture} presents the parallel RGB and optical-flow encoders and explains how their outputs are kept spatially and channel-wise aligned for downstream integration. We then describe the \textit{Flow-Guided Attention Mechanism}, which uses motion evidence from optical flow to refine RGB features through channel and spatial reweighting. After that, \textit{Empirical Mode Decomposition for Temporal Analysis} explains how the motion descriptor time series are decomposed into intrinsic mode functions and converted into a fixed-dimensional embedding. Finally, \textit{Temporal Modeling and Feature Fusion} and \textit{Classification and Training Objective} describe the bidirectional LSTM temporal encoding, multimodal fusion strategy, and the supervised training objective used to produce the final congestion state predictions.

\begin{table}[h!]
\centering
\caption{Mathematical notation used throughout the methodology section.}
\label{tab:notation}
\resizebox{12.5cm}{!}{
\begin{tabular}{ll}
\hline
\textbf{Symbol} & \textbf{Definition} \\
\hline
$\mathbf{V}=\{I_1,\ldots,I_T\}$ & Traffic video clip of length $T$ \\
$I_t\in\mathbb{R}^{H\times W\times C}$ & RGB frame at time $t$ \\
$O_t(x,y)$ & Optical flow vector at pixel $(x,y)$ between $I_t$ and $I_{t+1}$ \\
$u_t(x,y),\, v_t(x,y)$ & Horizontal and vertical components of $O_t(x,y)$ \\
$M_t(x,y),\, D_t(x,y)$ & Flow magnitude and flow direction at pixel $(x,y)$ \\
$m_t=[\mu(M_t),\,\sigma(M_t)]$ & Frame-level magnitude descriptors: spatial mean and spatial std of $M_t(x,y)$ \\
$d_t=[\mu(D_t),\,\sigma(D_t)]$ & Frame-level direction descriptors: spatial mean and spatial std of $D_t(x,y)$ \\
$\mathbf{X}_{rgb}^{(t)},\,\mathbf{X}_{flow}^{(t)}$ & RGB and flow feature maps at time $t$ \\
$\mathbf{A}_c,\,\mathbf{A}_s$ & Channel and spatial attention weights \\
$\tilde{\mathbf{X}}_{rgb}^{(t)}$ & Flow-enhanced RGB feature map after attention \\
$\mathbf{f}_{rgb},\,\mathbf{f}_{flow}$ & Clip-level temporal encodings from RGB and flow streams \\
$\mathbf{f}_{visual}$ & Fused visual representation from RGB and flow streams \\
$\mathbf{z}_{EMD}^{proc}$ & EMD-derived motion embedding for the clip \\
$\mathbf{h}_{combined}$ & Final fused feature vector used for classification \\
$\mathbf{y}_{pred}$ & Predicted class probability vector \\
$B,\,T,\,K$ & Batch size, clip length, number of classes \\
$H_f,\,W_f$ & Spatial dimensions of intermediate feature maps \\
\hline
\end{tabular}}
\end{table}

\subsection{Optical Flow Extraction and Motion Characterization}\label{sec:optical_flow}

The accurate representation of motion is important when it comes to distinguishing between free-flow and congested states, since spatial features alone may be misleading. Dense optical flow is estimated between consecutive frames using the Farneback algorithm. For each pixel $(x,y)$, the displacement vector is expressed as

\begin{equation}
O_t(x,y) = \begin{bmatrix} u_t(x,y) \\ v_t(x,y) \end{bmatrix},
\end{equation}

where $u_t(x,y)$ and $v_t(x,y)$ denote horizontal and vertical displacements between frames $I_t$ and $I_{t+1}$. From these, motion magnitude and direction are derived:

\begin{equation}
M_t(x,y) = \sqrt{u_t(x,y)^2 + v_t(x,y)^2}
\end{equation}
\begin{equation}
D_t(x,y) = \arctan2(v_t(x,y), u_t(x,y))
\end{equation}
The motion magnitude represents the Euclidean norm of the pixel displacement vector, quantifying the overall intensity of motion at location $(x,y)$, while direction captures its orientation. To extract compact statistical representations of frame-level motion patterns, spatial aggregation is performed over the pixel-wise motion fields to derive meaningful traffic descriptors. For each frame $t$, the motion magnitude field $M_t(x,y)$ and direction field $D_t(x,y)$ are summarized through first and second-order statistical moments computed across all spatial locations. Specifically, frame-level motion descriptors are defined as $\mathbf{m}_t = [\mu_{M_t}, \sigma_{M_t}]$ and $\mathbf{d}_t = [\mu_{D_t}, \sigma_{D_t}]$, where $\mu_{M_t}$ and $\mu_{D_t}$ represent the spatial mean of motion magnitude and direction respectively, while $\sigma_{M_t}$ and $\sigma_{D_t}$ capture their corresponding spatial standard deviations.

Our reason for using these statistical descriptors is motivated by their ability to characterize distinct traffic flow patterns. The mean magnitude $\mu_{M_t}$ quantifies the overall intensity of vehicular movement within the frame, with higher values indicating faster traffic flow, while the magnitude variance $\sigma_{M_t}$ captures the spatial heterogeneity of motion, reflecting the presence of mixed flow conditions such as stop-and-go traffic or lane-specific congestion patterns. Similarly, the directional statistics encode the coherence of traffic flow, where low directional variance indicates uniform flow direction typical of free-flowing conditions, while high variance suggests erratic or conflicting movement patterns characteristic of congested states. These compact descriptors serve as input to the EMD-based temporal analysis module, enabling the extraction of intrinsic oscillatory patterns in traffic behavior that complement the learned spatio-temporal features from the dual-stream architecture.

\subsection{Dual‐Stream Backbone Architecture}\label{sec:dual_stream}

To capture both spatial appearance and motion dynamics in a coherent and complementary manner, the framework employs a dual‐stream backbone composed of an RGB backbone and an optical‐flow backbone that operate in parallel and produce temporally aligned feature maps suitable for downstream attention, temporal encoding, and fusion. Given a mini‐batch of $B$ video clips with $T$ frames each, the RGB input has shape $B \times T \times 3 \times H \times W$ and the dense optical flow sequence (computed between consecutive frames) has shape $B \times (T-1) \times H \times W \times 2$. At time $t$, the streams yield feature maps
\begin{equation}
\mathbf{X}_{rgb}^{(t)} = f_{rgb}(I_t) \in \mathbb{R}^{C \times H' \times W'}, 
\qquad
\mathbf{X}_{flow}^{(t)} = f_{flow}(O_t) \in \mathbb{R}^{C \times H' \times W'},
\end{equation}
where $f_{rgb}$ and $f_{flow}$ are convolutional encoders with matched stage depths so that $\mathbf{X}_{rgb}^{(t)}$ and $\mathbf{X}_{flow}^{(t)}$ remain spatially and channel‐wise compatible ($C$ channels at resolution $H'\times W'$). This compatibility is essential for the flow‐guided attention module, which consumes both maps at the same spatial stride, and for the later fusion stage.

\begin{figure}[htbp]
\centering
\includegraphics[width=\textwidth]{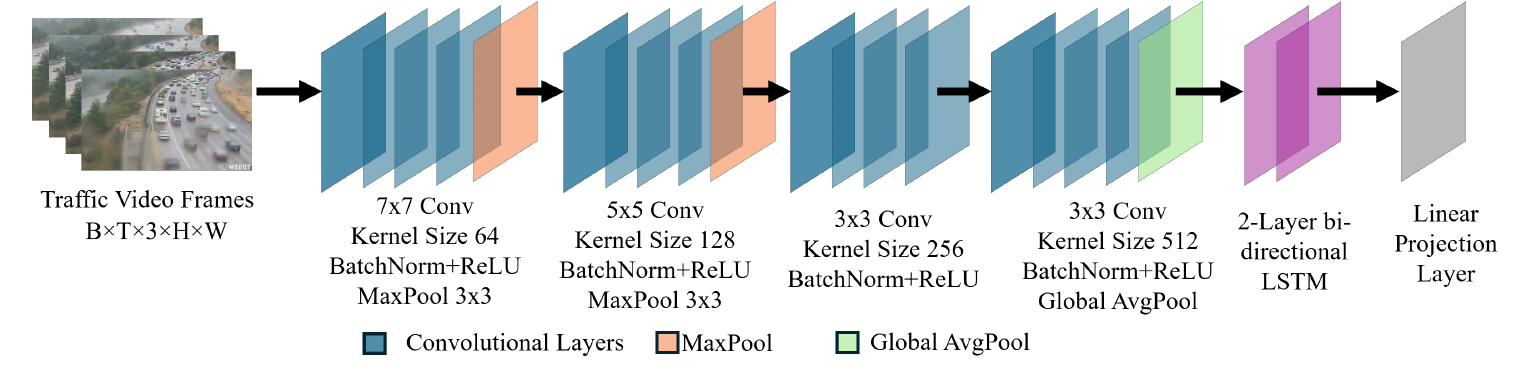}
\caption{RGB backbone: hierarchical convolutions (64, 128, 256, 512 channels) with decreasing spatial resolution, followed by global average pooling to obtain a fixed-dimensional frame representation.}
\label{fig:rgb_backbone}
\end{figure}

\paragraph*{RGB backbone:}
The RGB backbone (Figure~\ref{fig:rgb_backbone}) processes input tensors of shape $B \times T \times 3 \times H \times W$. It begins with a $7 \times 7$ convolution with 64 channels, batch normalization, and ReLU to stabilize early feature extraction, followed by $3 \times 3$ max pooling to reduce spatial size while retaining salient responses. Deeper blocks use progressively smaller kernels ($5 \times 5$, then $3 \times 3$) with increasing channels (128, 256, 512), which shifts the representation from edges and textures to object and scene-level cues such as vehicle clusters and road structure. A global average pooling layer converts the final feature map to a fixed-dimensional vector per frame, making the representation invariant to input resolution and ready for temporal encoding.
\paragraph*{Optical–flow backbone:}
This mirrors the RGB backbone design but operates on dense flow fields of shape $B \times (T{-}1) \times H \times W \times 2$. The first convolution adapts to the two-channel input $(u,v)$ and is followed by the same sequence of blocks and activations, ensuring that learned motion features (e.g., coherent motion regions, velocity changes, and directional patterns) are produced at the same spatial scales and channel dimensions as the RGB stream. As in the RGB backbone, a final global average pooling yields a fixed-length vector per frame (or flow step), aligned with the RGB embedding for downstream modules.

Keeping the two streams architecturally symmetric simplifies cross-modal alignment and reduces the need for additional projection layers. The RGB backbone preserves spatial appearance cues that help disambiguate scenes with similar motion but different layouts, while the flow backbone provides direct evidence of dynamic behavior. Their outputs feed the flow-guided attention mechanism, where the motion stream conditions the spatial and channel emphasis of the RGB features. The aligned, pooled vectors from both streams are then passed to the temporal encoder and the fusion module, ensuring that subsequent stages receive complementary and comparable representations.


\subsection{Flow-Guided Attention Mechanism}

\begin{figure}[htbp]
\centering
\includegraphics[width=0.8\textwidth]{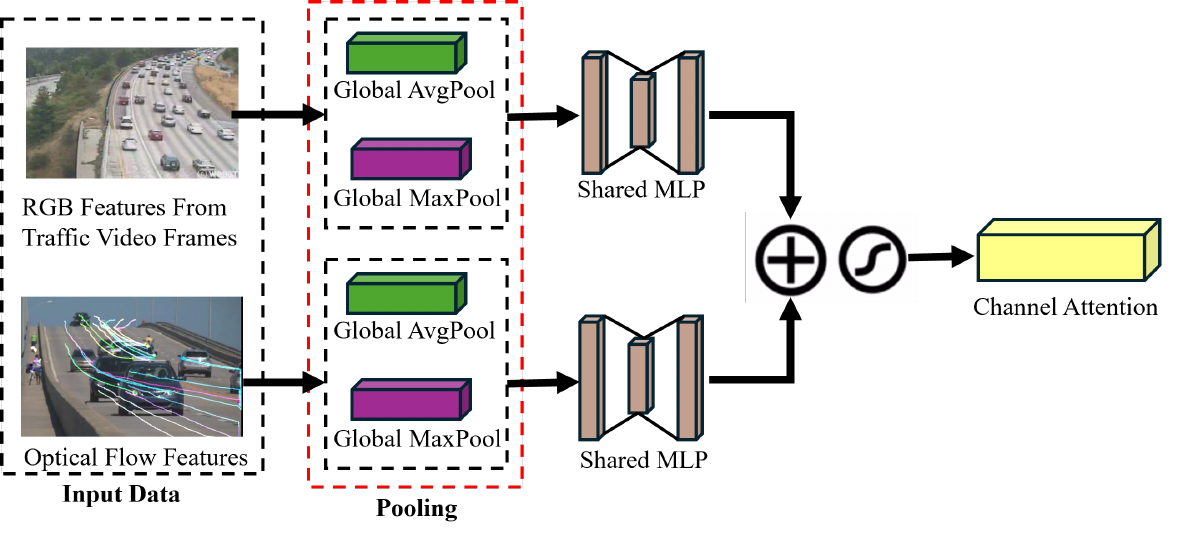}
\caption{Flow-guided channel attention mechanism. RGB and optical flow features undergo global average and maximum pooling, followed by processing through shared MLPs to generate channel attention weights that emphasize motion-relevant feature channels for traffic analysis.}
\label{fig:channel_attention}
\end{figure}

\begin{figure}[htbp]
\centering
\includegraphics[width=0.8\textwidth]{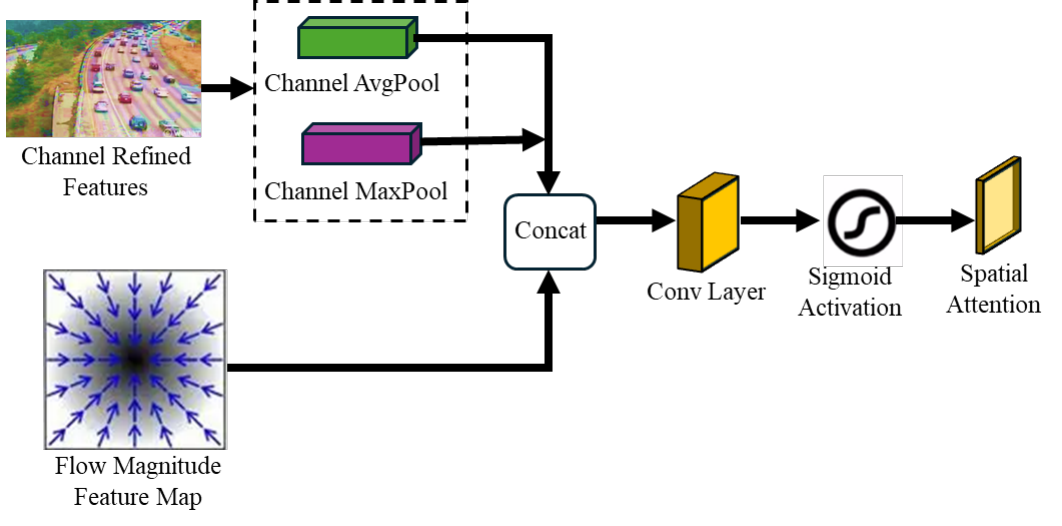}
\caption{Flow-guided spatial attention mechanism. Channel-refined RGB features are combined with flow magnitude information through channel-wise pooling operations. The concatenated spatial maps undergo convolution and sigmoid activation to produce spatial attention weights that focus on motion-active regions while suppressing static background elements.}
\label{fig:spatial_attention}
\end{figure}

The flow-guided attention module conditions RGB feature refinement on motion evidence, reducing reliance on appearance-only cues. The module follows the CBAM structure but introduces two modifications: (i) channel attention is computed from pooled descriptors of both RGB and flow features, and (ii) spatial attention incorporates a flow-magnitude map as an additional spatial cue. Figures~\ref{fig:channel_attention} and \ref{fig:spatial_attention} illustrate the channel and spatial attention modules.

\paragraph*{Channel attention.}
Given feature maps $\mathbf{X}_{rgb}^{(t)}$ and $\mathbf{X}_{flow}^{(t)}$, we first align the flow channels to the RGB channels using a $1\times 1$ convolution, then apply global average pooling and global max pooling to each modality. The pooled descriptors from RGB and flow are concatenated and passed through a shared MLP with reduction ratio $r=16$ to produce channel weights $\mathbf{A}_c\in\mathbb{R}^{B\times C\times 1\times 1}$. The RGB feature map is reweighted as $\mathbf{X}'^{(t)}_{rgb}=\mathbf{A}_c\odot \mathbf{X}^{(t)}_{rgb}$.

\paragraph*{Spatial attention.}
Spatial attention is computed from three spatial maps: the channel-wise mean and channel-wise max of $\mathbf{X}'^{(t)}_{rgb}$, and a resized, normalized flow-magnitude map derived from $M_t(x,y)$. These three maps are concatenated and processed with a $7\times 7$ convolution followed by a sigmoid to produce spatial weights $\mathbf{A}_s\in\mathbb{R}^{B\times 1\times H_f\times W_f}$. The final flow-enhanced RGB feature map is
\begin{equation}
\tilde{\mathbf{X}}_{rgb}^{(t)}=\mathbf{A}_s\odot \mathbf{X}'^{(t)}_{rgb}
= \mathbf{A}_s\odot\big(\mathbf{A}_c\odot \mathbf{X}^{(t)}_{rgb}\big).
\end{equation}

This design ties attention to motion-active regions while retaining appearance context, thereby improving robustness and interpretability in traffic scenes where static background elements may otherwise dominate attention.

\subsection{Temporal Modeling and Feature Fusion}

Temporal modeling integrates the enhanced spatial features across time to capture congestion formation and dissipation patterns. For each time step, $\tilde{\mathbf{X}}_{rgb}^{(t)}$ and $\mathbf{X}_{flow}^{(t)}$ are globally average pooled to obtain 512-dimensional vectors. Stacking over time yields an RGB sequence of shape $(B,T,512)$ and a flow sequence of shape $(B,T-1,512)$. To align temporal lengths, the flow sequence is padded with a zero vector at the final time step, yielding a $(B,T,512)$ sequence.

Separate two-layer bidirectional LSTMs encode the RGB and flow sequences using hidden size 256 and dropout 0.3. The BiLSTMs produce temporally contextualized representations for each time step, which are projected to 512 dimensions to maintain a consistent feature size across modules. To obtain clip-level descriptors, we apply attention-based temporal pooling. A small MLP produces an importance score for each time step, scores are normalized with a softmax over time, and the weighted sum yields clip embeddings $\mathbf{f}_{rgb}\in\mathbb{R}^{B\times 512}$ and $\mathbf{f}_{flow}\in\mathbb{R}^{B\times 512}$. These are concatenated and passed through a fusion MLP to form the unified visual representation $\mathbf{f}_{visual}\in\mathbb{R}^{B\times 512}$.

In parallel to the learned visual pathway, FLO-EMD constructs an explicit motion embedding using Empirical Mode Decomposition applied to compact optical-flow motion traces. From the optical-flow field, each frame contributes four scalar descriptors: mean flow magnitude, standard deviation of flow magnitude, mean flow direction, and standard deviation of flow direction. Over a clip, each descriptor forms a one-dimensional time series of length $T$, denoted $s^{(q)}=\{s^{(q)}_1,\ldots,s^{(q)}_T\}$ for descriptor index $q\in\{1,2,3,4\}$. EMD is applied independently to each $s^{(q)}$ to obtain a fixed number of intrinsic mode functions (IMFs),
\begin{equation}
s^{(q)}(t) = \sum_{j=1}^{N}\mathrm{IMF}^{(q)}_{j}(t) + r^{(q)}(t),
\end{equation}
where $N$ is the number of retained IMFs and $r^{(q)}(t)$ is the residual. In this work, we retain $N=4$ IMFs per descriptor time series.

To convert the IMFs into a fixed-dimensional representation, each $\mathrm{IMF}^{(q)}_{j}(t)$ is summarized by two temporal statistics, its temporal mean and temporal standard deviation. This yields $4$ descriptors $\times$ $4$ IMFs $\times$ $2$ statistics $=32$ scalar features per clip, which are flattened and passed through a dedicated projection MLP to obtain the processed EMD embedding:
\begin{equation}
\mathbf{z}^{proc}_{EMD} = \mathrm{MLP}_{EMD}\big(\mathrm{Flatten}(\mathbf{z}_{IMF})\big)\in\mathbb{R}^{B\times 128}.
\end{equation}
This branch provides a temporally adaptive, multi-scale characterization of traffic motion that complements the learned spatiotemporal features.

The final representation is formed by concatenating the fused visual embedding with the EMD embedding:
\begin{equation}
\mathbf{h}_{combined}=[\mathbf{f}_{visual};\mathbf{z}^{proc}_{EMD}]\in\mathbb{R}^{B\times 640},
\end{equation}
so that the classifier leverages both learned appearance-motion cues and EMD-derived multi-scale motion components.

\subsection{Classification and Training Objective}

The classification head maps the fused representation $\mathbf{h}_{combined}$ to congestion state probabilities using a three-layer MLP with progressive dimensionality reduction $(640\rightarrow 512\rightarrow 256\rightarrow K)$, where $K=3$. Each hidden layer uses ReLU activation and dropout (0.5, 0.3, 0.3) for regularization. The final layer outputs logits, which are converted to probabilities via softmax at inference.

Training uses cross-entropy loss with label smoothing to reduce overconfidence and improve generalization in borderline cases between adjacent congestion levels:
\begin{equation}
L = -\frac{1}{|B|}\sum_{i=1}^{|B|}\sum_{k=1}^{K} \tilde{y}_{i,k}\log(p_{i,k}),
\qquad
\tilde{y}_{i,k}=(1-\epsilon)y_{i,k}+\frac{\epsilon}{K},
\end{equation}
where $p_{i,k}$ is the predicted probability and $\tilde{y}_{i,k}$ is the smoothed target.

For interpretability, the spatial attention maps $\mathbf{A}_s$ can be visualized across time to verify that the model emphasizes motion-relevant regions such as queue formations and lane groups. During optimization, all learnable network components are trained jointly under the supervised objective, while the EMD branch provides complementary multi-scale temporal motion components computed from the optical-flow motion traces.

\subsection{Implementation Details}
All video frames are resized to $224 \times 224$, and each clip is represented using $T=16$ uniformly sampled frames. The RGB and optical-flow backbones share the same stage design with channel widths of 64, 128, 256, and 512, producing spatially aligned feature maps for multimodal fusion. The attention module follows a channel--spatial design, using a channel reduction ratio of $r=16$ and a $7 \times 7$ convolution for spatial attention. Temporal modeling is implemented using a two-layer bidirectional LSTM with hidden size 256 and dropout 0.3, followed by attention-based temporal pooling.

For the EMD branch, each aggregated motion-trace time series is decomposed into $N=4$ intrinsic mode functions. Each intrinsic mode function is summarized by its temporal mean and standard deviation, and the resulting fixed-length vector is mapped to a 128-dimensional embedding using a multilayer perceptron. The classification head is a multilayer perceptron with hidden layer sizes 512 and 256 and an output layer of size 3, with dropout rates of 0.5, 0.3, and 0.3 applied across successive layers.

\section{Experimental Setup}
All experiments were conducted using an NVIDIA L4 GPU (20GB VRAM) with 32GB system memory, implemented in Python 3.9 and PyTorch 2.0. This section details the experimental configuration for training and validating our flow-guided congestion classification framework for traffic video analysis, including the model settings, optimization strategy, and evaluation protocol used to ensure reproducible results.

\subsection{Dataset}

\begin{figure*}[htbp]
\centering
\includegraphics[width=0.85\textwidth]{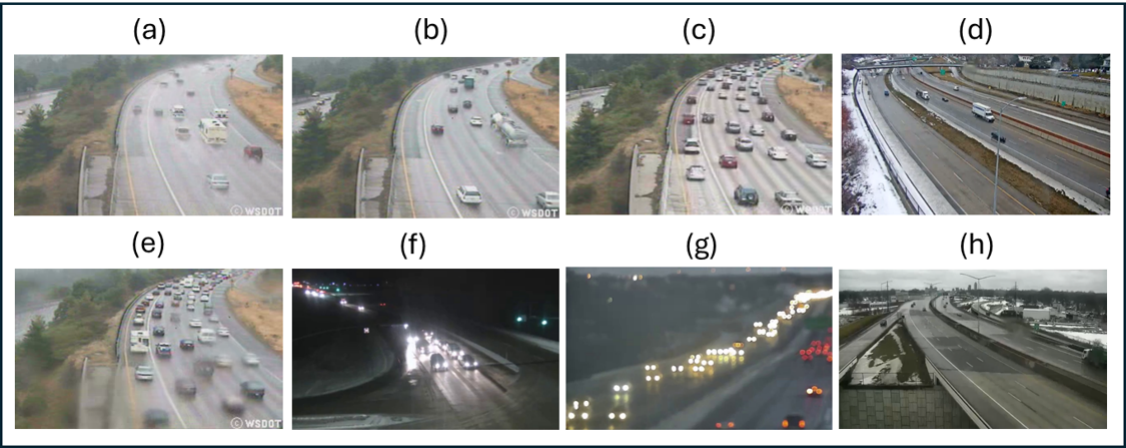}
\caption{Dataset diversity showing various traffic conditions and environmental scenarios: (a) light traffic flow on a multi-lane highway under clear daytime conditions, (b) moderate traffic density with well-spaced vehicles during clear weather, (c) increased traffic density showing more congested but still flowing conditions, (d) highway infrastructure with moderate traffic under overcast conditions, (e) moderate traffic density on a multi-lane highway during rainy daytime period, (f) nighttime traffic scene with visible vehicle headlights and taillights, (g) nighttime highway conditions showing illuminated vehicle lights in low visibility, and (h) complex highway interchange system with light traffic flow. The dataset encompasses multiple lighting conditions, traffic densities, and road configurations to ensure comprehensive model evaluation across diverse scenarios.}
\label{fig:dataset_diversity}
\end{figure*}

This study employs a multi-source traffic surveillance dataset compiled to support evaluation across diverse traffic conditions and environmental scenarios (Figure \ref{fig:dataset_diversity}). The dataset integrates footage from four complementary sources: (1) 254 highway clips from Interstate 5 in Seattle, Washington, originally compiled by the Statistical Visual Computing Lab at the University of California, San Diego \cite{chan2005probabilistic}; (2) selected clips from the AI-City Challenge 2021 Track 4 dataset captured at multiple intersections and highways in Iowa, USA \cite{naphade2021aicity}; (3) urban arterial and freeway footage from the Tennessee Department of Transportation (TDOT) SmartWay traffic management system \cite{TDOTSmartWay}; and (4) expressway and bridge sequences from the NYC DOT 511NY surveillance network \cite{511NY}. Collectively, these sources provide representative coverage of North American monitoring contexts, including highway, arterial, intersection, bridge, and dense urban environments under varied weather and lighting.

All videos are standardized to 5-second clips at 30 fps, with resolution normalized to $224\times224$ during preprocessing. The complete dataset contains 1,050 clips, and each clip is categorized by observable environmental conditions through manual inspection (as summarized in Table~\ref{tab:dataset_composition}).

\begin{table}[htbp]
\caption{Dataset composition by environmental and weather conditions.}
\label{tab:dataset_composition}
\centering
\small
\begin{tabular}{lccccc}
\hline
\textbf{Condition} & \textbf{Train} & \textbf{Val} & \textbf{Test} & \textbf{Total} \\
\hline
Clear Daytime & 260 & 30 & 60 & 350\\
Nighttime (Clear) & 110 & 15 & 25 & 150\\
Overcast/Cloudy & 105 & 12 & 23 & 140\\
Rainy Conditions & 95 & 10 & 25 & 130\\
Winter/Snow & 90 & 10 & 20 & 120\\
Fog/Low Visibility & 105 & 8 & 47 & 160\\
\hline
\textbf{Total} & \textbf{765} & \textbf{85} & \textbf{200} & \textbf{1,050} \\
\hline
\end{tabular}
\end{table}

Traffic congestion labels are manually assigned through visual inspection following the qualitative labeling protocol established in the UCSD traffic dataset \cite{chan2005probabilistic}. Each clip is classified into one of three congestion states based on prevailing speed regime and flow characteristics: \textit{light} denotes free-flow conditions with ample spacing and stable movement, \textit{medium} denotes reduced speeds with tighter spacing but sustained motion, and \textit{heavy} denotes stop-and-go or near-stationary traffic with dense queues. Labeling was performed independently by two annotators with transportation engineering background, with disagreements resolved through consensus review.

To prevent data leakage and to evaluate generalization, partitioning was performed at the scene (camera viewpoint) level such that no camera location appears in more than one split. The training, validation, and test sets contain 765, 85, and 200 clips, respectively. The resulting class distributions are: training (312 light, 268 medium, 185 heavy), validation (32 light, 31 medium, 22 heavy), and test (68 light, 78 medium, 54 heavy). This protocol ensures that performance reflects the model's ability to generalize to unseen camera viewpoints and operational conditions.

\subsection{Loss Function and Training Procedure}
All learnable network components are trained jointly using cross-entropy loss with label smoothing ($\epsilon=0.1$) to reduce overconfident predictions. Optimization is performed with Adam (initial learning rate $1\times10^{-3}$) using a step-decay schedule that multiplies the learning rate by 0.1 at epochs 30 and 60. Models are trained for 300 epochs with batch size 4, and gradients are clipped to a maximum $\ell_2$ norm of 1.0 to stabilize training. Dropout is applied in the classification head (0.5, 0.3, 0.3). To preserve alignment between RGB frames and optical flow inputs, spatial augmentations are not used.

\subsection{Baseline Comparisons and Competitive Analysis}
We compare FLO-EMD against representative video-recognition and hybrid baselines adapted to three-class congestion classification under a unified evaluation protocol. The baselines are grouped into three families: transformer-based video models, CNN-based video models, and hybrid pipelines that combine per-frame perception with explicit temporal modeling. Specifically, we evaluate TimeSformer \cite{bertasius2021space}, Video Swin Transformer \cite{liu2022video}, and MViTv2 \cite{li2022mvitv2} as transformer baselines; X3D \cite{feichtenhofer2020x3d}, TANet \cite{liu2021tam}, TEA \cite{li2020tea}, TSM \cite{lin2019tsm} as CNN baselines; and detection-plus-recurrence pipelines (YOLO+LSTM \cite{s25216699} and YOLO+GRU \cite{yulin2025research}) as hybrid baselines.

For fairness, each baseline retains its original backbone while replacing the classifier with a three-way output head (light, medium, heavy). All methods use the same input preprocessing and clip formation (16 frames per clip at $224\times224$). CNN and hybrid baselines are trained from scratch on the proposed dataset using the same optimizer, learning-rate schedule, number of epochs, batch size, and regularization settings. Spatial augmentations are not applied to preserve correspondence between RGB frames and optical flow inputs. This protocol ensures that performance differences primarily reflect architectural design choices rather than pretraining or training variability.

All transformer baselines (TimeSformer, Video Swin Transformer, and MViTv2) were initialized from Kinetics-400 pretrained weights and fine-tuned end-to-end on the proposed congestion dataset, with all parameters trainable and the original classification layers replaced by a three-way classifier head (light, medium, heavy) for direct comparability. Each input clip comprises 16 frames uniformly sampled with a temporal stride of 4 and resized to $224\times224$, and we do not apply spatial augmentations to preserve RGB-flow correspondence. Fine-tuning uses AdamW with learning rate $1\times10^{-4}$ and a 5-epoch linear warm-up, followed by cosine annealing for the remainder of training; weight decay is 0.05, drop-path rate is 0.15, batch size is 4, and training runs for 300 epochs. Gradients are clipped with a maximum norm of 1.0 to stabilize optimization, and testing uses single-clip, center-crop evaluation

\subsection{Robustness Across Environmental Conditions}
To assess deployment robustness, we report FLO-EMD performance on the held-out test set stratified by environmental condition. Test clips are grouped into six visually identified categories: clear daytime (60 clips), fog/low visibility (47 clips), nighttime (25 clips), rainy (25 clips), overcast/cloudy (23 clips), and winter/snow (20 clips). For each category, we report accuracy, weighted F1, and cross-entropy loss, enabling direct comparison of performance under changes in illumination, visibility, and precipitation.

\subsection{Component Analysis and Interpretability}
We perform controlled studies to isolate the contribution of each FLO-EMD component and to assess whether the attention mechanism yields operationally meaningful explanations. First, an ablation study evaluates progressively simplified variants of the model by removing flow-guided attention, the EMD branch, and temporal modeling while keeping all other settings fixed. Second, an EMD sensitivity analysis varies the number of intrinsic mode functions extracted per motion-trace time series and evaluates performance stability, identifying a parsimonious configuration. Third, interpretability is examined through attention visualizations overlaid on video frames to verify whether high-attention regions align with traffic-relevant motion patterns across congestion levels and environmental conditions.

\section{Evaluation Metrics}
We evaluate three-class congestion classification using accuracy and class-aware F1 scores. In addition to overall accuracy, we report macro-averaged precision, recall, and F1 to weight all classes equally, and weighted-averaged precision, recall, and F1 to account for class imbalance via class support. We also report cross-entropy loss and present confusion matrices to characterize misclassification patterns.

\section{Results and Discussion}
This section presents quantitative and qualitative evidence supporting the proposed FLO-EMD framework. We first compare FLO-EMD against representative CNN-based video models, transformer-based video models, and hybrid detection-plus-recurrence baselines under a unified protocol. We then isolate the contribution of each component through ablation and EMD sensitivity analyses. Finally, we evaluate robustness across environmental conditions and discuss interpretability results from attention visualizations.

\subsection{Comparison with Baselines Across Model Families}
Table~\ref{tab:comprehensive_results} summarizes test-set performance across all baselines and the proposed method. FLO-EMD achieves the highest overall accuracy (97.50\%) and the lowest cross-entropy loss (0.0789), with a weighted F1 of 0.9742. Class-specific F1 scores remain consistently high across congestion states (Light: 0.9808, Medium: 0.9538, Heavy: 0.9721), indicating balanced sensitivity.

To enable clearer cross-comparison, the baselines are discussed by architectural family.

\paragraph*{CNN baselines.}
Among CNN-based video models, X3D and TANet provide the strongest performance (91.00\% and 89.50\% accuracy, respectively), reflecting the utility of spatiotemporal convolutions for capturing local motion patterns. However, these models remain largely appearance-driven at the feature-selection stage. In traffic scenes, high-contrast but static elements can correlate with congestion labels without being causal indicators of traffic state. FLO-EMD improves upon these CNN baselines by explicitly conditioning feature emphasis on motion evidence through optical-flow guidance, which reduces reliance on static infrastructure cues and improves class balance.

\paragraph*{Hybrid pipelines.}
Hybrid pipelines achieve stronger results than most single-family baselines, with TSM providing the strongest baseline overall (95.50\% accuracy; weighted F1 = 0.9534). This indicates that explicit temporal interaction can improve congestion recognition when combined with a strong spatial backbone. Nevertheless, these pipelines do not directly address a key reliability and interpretability failure mode in surveillance video: feature importance can still concentrate on visually salient but traffic-irrelevant regions. FLO-EMD improves on these baselines by introducing motion-conditioned spatial and channel refinement and by augmenting learned video features with data-adaptive temporal decomposition of motion traces, yielding higher accuracy and substantially lower loss.

\paragraph*{Transformer baselines and spatial misalignment.}
Transformer-based architectures perform substantially well with full fine-tuning from Kinetics-400 pretrained weights, with accuracies ranging from 90.50\% to 94.00\% and Video Swin providing the strongest transformer baseline (94.00\% accuracy; weighted F1 = 0.9370). Despite these gains, the transformer baselines still underperform FLO-EMD, which is consistent with two factors that are particularly relevant to congestion classification.

First, standard transformer attention is computed over a fixed token grid derived from RGB appearance. While self-attention can discover long-range relations, it does not inherently enforce alignment between attended regions and motion-relevant evidence. In traffic surveillance footage, the most visually salient regions are often static infrastructure elements. As a result, attention weights can drift toward shoulders, lane boundaries, and road texture patterns that remain stable across time, producing spatial misalignment between attended regions and the causal motion signals that distinguish free-flow from congestion.

Second, several video transformers factorize attention across space and time or restrict attention to local windows to control computation. These design choices can further weaken the model's ability to track subtle changes in motion structure over time, particularly when congestion states differ primarily in velocity regimes and stop-and-go dynamics rather than in distinct object categories. Under limited domain-specific training data and without strong inductive biases toward motion, these transformers can over-attend to static cues and under-utilize temporal motion evidence.

FLO-EMD directly addresses this spatial misalignment through flow-guided attention. Optical flow provides explicit motion evidence, and the proposed attention module uses flow features and flow magnitude to modulate RGB channel and spatial weights. This conditioning mechanism encourages attention to follow traffic-active regions rather than static, high-contrast background structure, improving both performance and interpretability in surveillance settings.

\begin{table*}[h]
\centering
\caption{Comprehensive performance comparison of FLO-EMD with established baselines and hybrid architectures on the test set. Macro-averaged metrics treat all classes equally; weighted-averaged metrics account for class support. Best results are in bold. Transformer models marked with $\dagger$ indicate fine-tuning from Kinetics-400 pretrained weights.}
\label{tab:comprehensive_results}
\resizebox{13cm}{!}{
\small
\begin{tabular}{lccccccccccc}
\hline
\textbf{Model} & \textbf{Acc.} & \textbf{Loss} & \textbf{Macro} & \textbf{Macro} & \textbf{Macro} & \textbf{Wtd.} & \textbf{Wtd.} & \textbf{Wtd.} & \textbf{F1} & \textbf{F1} & \textbf{F1} \\
 & \textbf{(\%)} &  & \textbf{Prec.} & \textbf{Rec.} & \textbf{F1} & \textbf{Prec.} & \textbf{Rec.} & \textbf{F1} & \textbf{Light} & \textbf{Med.} & \textbf{Heavy} \\
\hline
\multicolumn{12}{c}{\textit{CNN Baselines}} \\
\hline
TANet~\cite{liu2021tam} & 89.50 & 0.2156 & 0.8812 & 0.8654 & 0.8721 & 0.8903 & 0.8950 & 0.8924 & 0.9423 & 0.8125 & 0.8615 \\
X3D~\cite{feichtenhofer2020x3d} & 91.00 & 0.2189 & 0.8245 & 0.8298 & 0.8256 & 0.8678 & 0.9100 & 0.8732 & 0.8654 & 0.7812 & 0.8302 \\
TEA~\cite{li2020tea} & 87.50 & 0.4521 & 0.7989 & 0.8124 & 0.8042 & 0.8512 & 0.8750 & 0.8598 & 0.9512 & 0.6875 & 0.7738 \\
\hline
\multicolumn{12}{c}{\textit{Transformer Baselines (Fine-tuned from Kinetics-400)}} \\
\hline
TimeSformer$^{\dagger}$~\cite{bertasius2021space} & 92.50 & 0.1834 & 0.9089 & 0.9012 & 0.9038 & 0.9178 & 0.9250 & 0.9211 & 0.9538 & 0.8623 & 0.8952 \\
Video Swin$^{\dagger}$~\cite{liu2022video} & 94.00 & 0.1523 & 0.9267 & 0.9189 & 0.9215 & 0.9345 & 0.9400 & 0.9370 & 0.9615 & 0.8912 & 0.9119 \\
MViTv2$^{\dagger}$~\cite{li2022mvitv2} & 90.50 & 0.2345 & 0.8834 & 0.8756 & 0.8782 & 0.8945 & 0.9050 & 0.8995 & 0.9385 & 0.8334 & 0.8627 \\
\hline
\multicolumn{12}{c}{\textit{Hybrid Pipelines}} \\
\hline
TSM~\cite{lin2019tsm} & 95.50 & 0.1234 & 0.9412 & 0.9398 & 0.9402 & 0.9523 & 0.9550 & 0.9534 & 0.9615 & 0.9154 & 0.9438 \\
YOLO+LSTM~\cite{s25216699} & 75.50 & 0.6834 & 0.7289 & 0.7154 & 0.7198 & 0.7456 & 0.7550 & 0.7489 & 0.8654 & 0.6589 & 0.6351 \\
YOLO+GRU-Attn~\cite{yulin2025research} & 81.50 & 0.5234 & 0.7989 & 0.7923 & 0.7941 & 0.8112 & 0.8150 & 0.8125 & 0.8923 & 0.7512 & 0.7388 \\
\hline
\multicolumn{12}{c}{\textit{Proposed Model}} \\
\hline
\textbf{FLO-EMD} & \textbf{97.50} & \textbf{0.0789} & \textbf{0.9689} & \textbf{0.9698} & \textbf{0.9689} & \textbf{0.9738} & \textbf{0.9750} & \textbf{0.9742} & \textbf{0.9808} & \textbf{0.9538} & \textbf{0.9721} \\
\hline
\end{tabular}}
\end{table*}

\subsection{Component Contribution and Cross-Comparison via Ablation}
Table~\ref{tab:ablation_study_results} reports ablation results designed to isolate the contribution of each component on the test set. The full FLO-EMD model achieves the best generalization profile, with the highest test accuracy (97.50\%) and the smallest train--test gap (2.01\%).

Several patterns are consistent and informative. Incorporating standard attention without motion guidance increases overfitting, as shown by the large generalization gap in FLO-EMD V2 (10.25\%). This supports the hypothesis that appearance-driven attention can lock onto visually prominent but non-causal scene elements in traffic footage. Adding optical flow without attention (FLO-EMD V3) improves motion awareness but remains incomplete because the model lacks an explicit mechanism to use motion to refine spatial emphasis within the RGB representation, and it exhibits reduced test performance (87.50\%) with a sizable gap (7.08\%).

The strongest partial variant is FLO-EMD V4, which includes optical flow and flow-guided attention but removes EMD analysis. Its performance (94.50\% test accuracy; 3.53\% gap) indicates that motion-conditioned feature refinement is the dominant driver of improvement, while the remaining gain from the full model quantifies the added value of adaptive temporal decomposition. FLO-EMD V5 confirms that temporal modeling remains necessary even with strong spatial refinement, because congestion is defined not only by spatial density but also by temporal evolution, including acceleration patterns, speed instability, and stop-and-go regimes (93.00\% test accuracy; 6.23\% gap).

\begin{table*}[ht]
    \centering
    \caption{Ablation Study}
    \label{tab:ablation_study_results}
    \resizebox{15cm}{!}{
    \begin{tabular}{lccccccc}
        \toprule
        \textbf{Model} & \textbf{Flow-Guided} & \textbf{EMD} & \textbf{Optical} & \textbf{Temporal} & \textbf{Train Acc.} & \textbf{Test Acc.} & \textbf{Gap}\\
        & \textbf{Attention} & \textbf{Analysis} & \textbf{Flow} & \textbf{Modeling} & (\%) & (\%) & (\%)\\
        \midrule
        FLO-EMD V1 (Baseline) & No & No & No & Yes & 99.02 & 93.50 & 5.52 \\
        FLO-EMD V2 & No & No & No & Yes & 99.75 & 89.50 & 10.25 \\
        FLO-EMD V3 & No & No & Yes & Yes & 94.58 & 87.50 & 7.08 \\
        FLO-EMD V4 & Yes & No & Yes & Yes & 98.03 & 94.50 & 3.53 \\
        FLO-EMD V5 & Yes & Yes & Yes & No & 99.23 & 93.00 & 6.23 \\
        FLO-EMD (Full model) & Yes & Yes & Yes & Yes & 99.51 & \textbf{97.50} & \textbf{2.01} \\
        \bottomrule
    \end{tabular}}
\end{table*}

\subsection{EMD Sensitivity and Motion-Descriptor Analysis}

Table~\ref{tab:imf_sensitivity} and Table~\ref{tab:motion_descriptor_ablation} summarize two controlled analyses that quantify the contribution of the EMD branch on identical data splits. First, the IMF sensitivity study shows that increasing the number of intrinsic mode functions improves generalization up to $N=4$ (97.50\% test accuracy, 2.01\% train-test gap). Performance is lower for $N=2$--3, indicating insufficient decomposition of distinct temporal behaviors, while $N\geq 5$ yields marginally lower test accuracy with a larger generalization gap, suggesting that additional modes begin to capture short-lived fluctuations that do not transfer across scenes.

Second, the motion-descriptor ablation indicates that aggregated magnitude statistics are the primary drivers of performance. Removing the magnitude mean ($\mu_M$) produces the largest drop (95.50\%, $-2.00$), consistent with $\mu_M$ representing overall motion intensity that separates free-flow from congested regimes. Removing magnitude variability ($\sigma_M$) also reduces performance (96.50\%, $-1.00$), reflecting the importance of heterogeneity in mixed and stop-and-go conditions. Directional statistics contribute complementary information (w/o $\mu_D$: 96.85\%, w/o $\sigma_D$: 97.15\%), and using only direction is insufficient (94.50\%). Overall, these results support using compact, aggregated optical-flow traces as stable non-stationary signals for EMD, and justify the choice $N=4$ in the full FLO-EMD configuration.

\begin{table}[t]
\centering
\caption{IMF sensitivity analysis for the EMD component.}
\label{tab:imf_sensitivity}
\small
\setlength{\tabcolsep}{8pt}
\begin{tabular}{lccc}
\hline
\textbf{Number of IMFs} & \textbf{Train Acc. (\%)} & \textbf{Test Acc. (\%)} & \textbf{Gap (\%)} \\
\hline
2 & 98.82 & 94.50 & 4.32 \\
3 & 99.15 & 95.50 & 3.65 \\
\textbf{4} & \textbf{99.51} & \textbf{97.50} & \textbf{2.01} \\
5 & 99.58 & 97.00 & 2.58 \\
6 & 99.73 & 96.50 & 3.23 \\
\hline
\end{tabular}
\end{table}

\begin{table}[t]
\centering
\caption{Motion descriptor ablation analysis for the EMD component.}
\label{tab:motion_descriptor_ablation}
\small
\setlength{\tabcolsep}{8pt}
\begin{tabular}{lcc}
\hline
\textbf{Descriptors Used} & \textbf{Test Acc. (\%)} & \textbf{Change in Test Acc. (\%)} \\
\hline
All four descriptors & \textbf{97.50} & \textbf{0.00} \\
Without $\mu_M$ & 95.50 & -2.00 \\
Without $\sigma_M$ & 96.50 & -1.00 \\
Without $\mu_D$ & 96.85 & -0.65 \\
Without $\sigma_D$ & 97.15 & -0.35 \\
Magnitude only & 96.00 & -1.50 \\
Direction only & 94.50 & -3.00 \\
\hline
\end{tabular}
\end{table}

\subsection{Robustness Under Diverse Environmental Conditions}

Table~\ref{tab:environmental_performance} summarizes FLO-EMD performance stratified by environmental condition on the held-out test set. The model remains robust across all scenarios, achieving at least 93.50\% accuracy in every category and an overall accuracy of 97.50\% (weighted F1 = 0.9742). Performance is strongest under clear daytime conditions (98.50\%), where visibility and motion boundaries are least corrupted, and remains consistently high under overcast (97.85\%) and fog/low-visibility conditions (96.75\%), indicating that motion-conditioned feature refinement can maintain discriminative cues even when contrast is reduced. Nighttime clips exhibit modest degradation (96.20\%), which is consistent with reduced illumination and headlight glare that can obscure spatial detail and introduce motion blur. Winter/snow conditions also show limited performance loss (95.50\%), suggesting that the model can still extract stable motion patterns despite texture changes and partial occlusions.

\begin{table}[h]
\centering
\caption{FLO-EMD performance across environmental conditions on expanded test set (200 samples).}
\label{tab:environmental_performance}
\small
\begin{tabular}{lcccc}
\hline
\textbf{Condition}  & \textbf{Accuracy (\%)} & \textbf{Weighted F1} & \textbf{Loss} \\
\hline
Clear Daytime  & 98.50 & 0.9846 & 0.0423 \\
Overcast/Cloudy  & 97.85 & 0.9781 & 0.0612 \\
Nighttime (Clear)  & 96.20 & 0.9615 & 0.0891 \\
Winter/Snow  & 95.50 & 0.9542 & 0.1123 \\
Fog/Low Visibility  & 96.75 & 0.9668 & 0.0834 \\
Rainy Conditions  & 93.50 & 0.9338 & 0.1567 \\
\hline
\textbf{Overall}  & \textbf{97.50} & \textbf{0.9742} & \textbf{0.0789} \\
\hline
\end{tabular}
\end{table}

Rainy conditions are the most challenging setting (93.50\% accuracy, weighted F1 = 0.9338), consistent with reduced visibility and appearance artifacts such as reflections and streaking that can degrade both RGB cues and motion estimation. Despite this degradation, performance remains high, indicating that motion-conditioned refinement and aggregated motion-trace features provide resilience when appearance quality is reduced.

\subsection{Qualitative and Interpretability Analysis}
\begin{figure}[ht]
\centering
\includegraphics[width=0.75\textwidth]{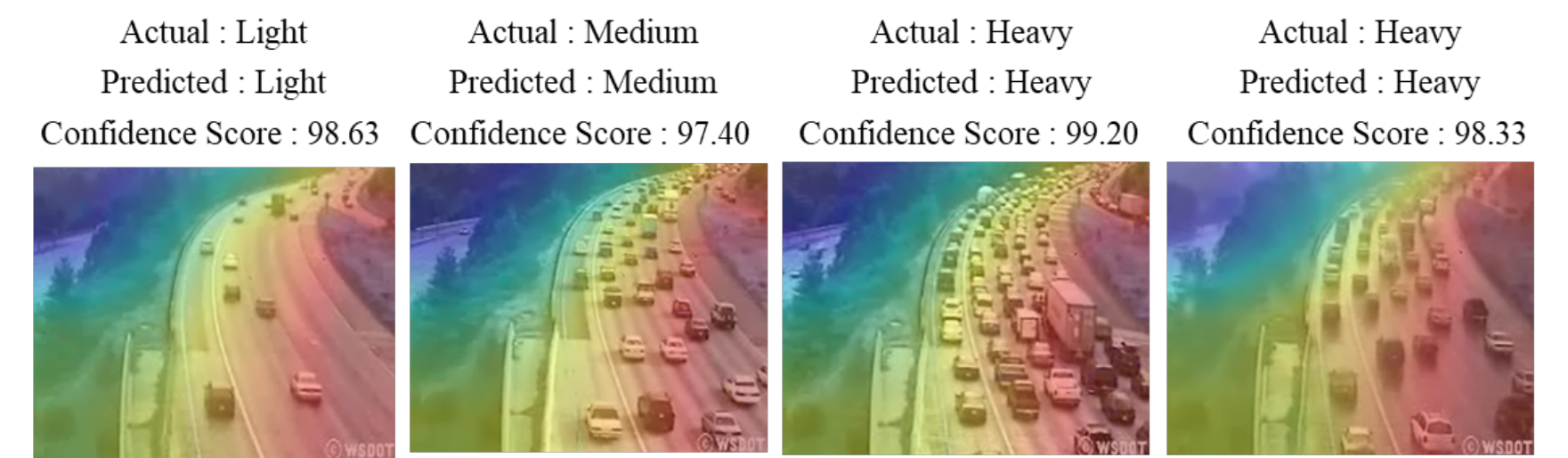}
\caption{Accurate classification examples showing the model's ability to correctly identify light, medium, and heavy traffic conditions with high confidence scores ranging from 97.40\% to 99.20\%. The attention heatmaps demonstrate consistent focus on traffic-relevant regions across different congestion levels.}
\label{fig:qualitative_results}
\end{figure}

\begin{figure}[ht]
\centering
\includegraphics[width=0.6\textwidth]{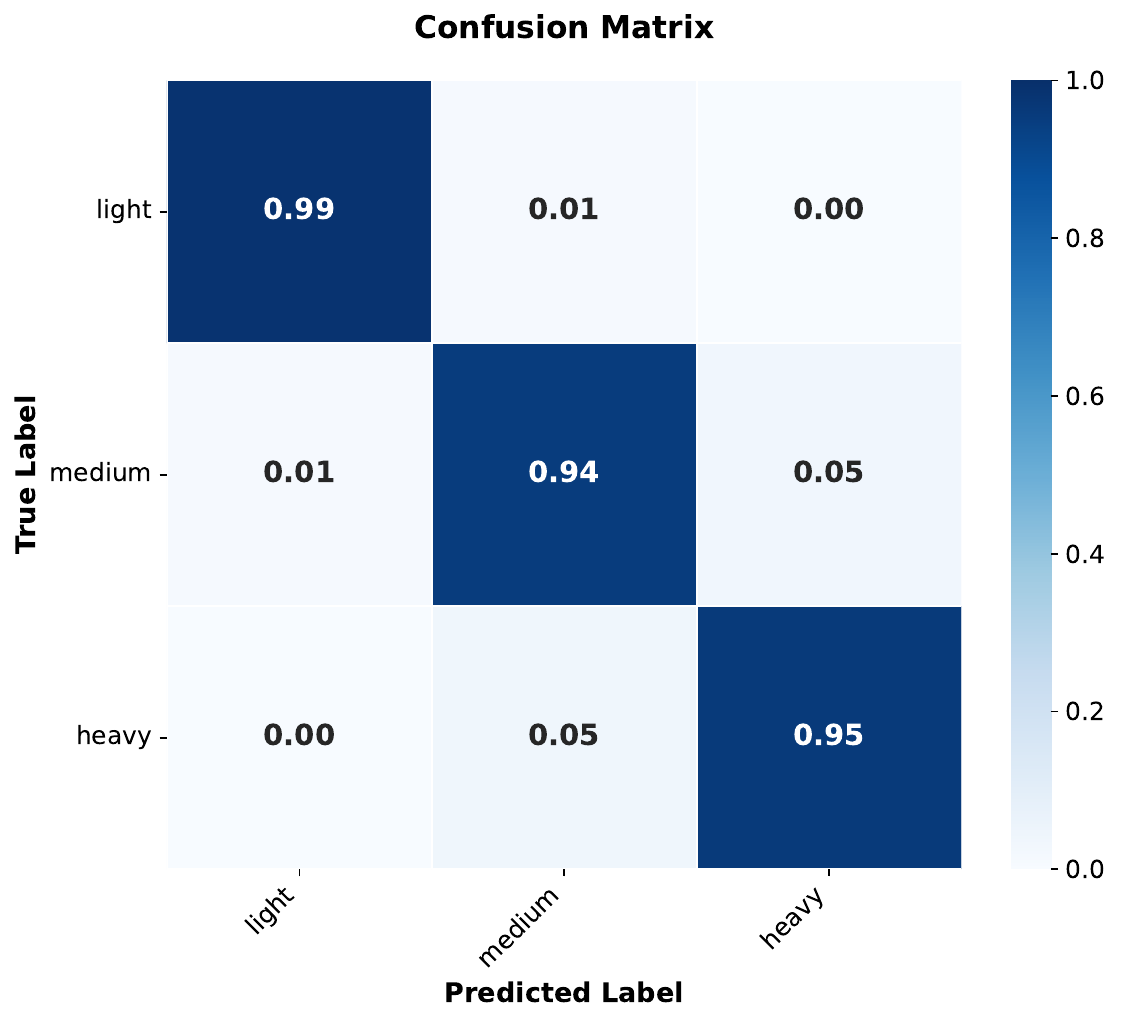}
\caption{Confusion matrix for the proposed model showing classification performance across all traffic congestion classes. The matrix reveals strong diagonal performance with minimal misclassification between adjacent congestion levels.}
\label{fig:confusion_matrix}
\end{figure}

\begin{figure}[ht]
\centering
\includegraphics[width=0.75\textwidth]{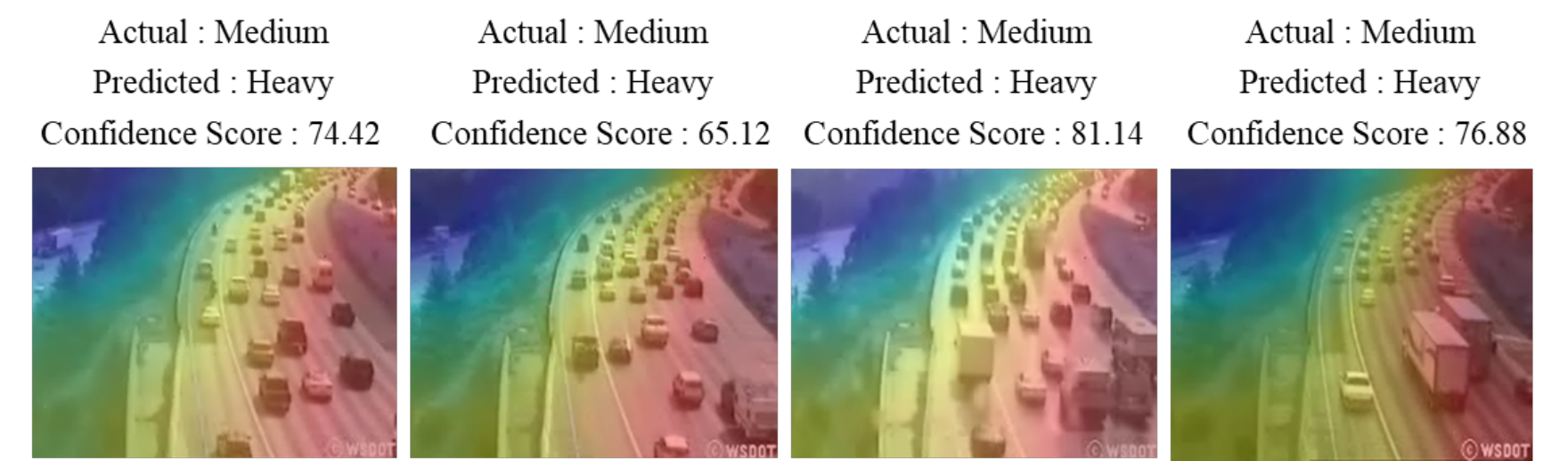}
\caption{Representative misclassification examples showing boundary cases where the model struggles with ambiguous traffic scenarios. The moderate confidence scores (65-81\%) indicate uncertainty in these challenging transition cases between medium and heavy congestion.}
\label{fig:qualitative_errors}
\end{figure}

Figures~\ref{fig:qualitative_results}-\ref{fig:qualitative_errors} provide qualitative evidence consistent with the quantitative metrics. Figure~\ref{fig:qualitative_results} shows correctly classified clips across all three congestion levels, where the model focuses on traffic-active regions and achieves high confidence (typically above 97\%). In contrast, Figure~\ref{fig:qualitative_errors} illustrates representative failure cases with lower confidence (approximately 65-81\%), indicating genuine ambiguity rather than systematic confusion.

Figure~\ref{fig:confusion_matrix} clarifies where the remaining errors occur. The matrix is strongly diagonal, with near-perfect recognition of light congestion at 99\% correct and high accuracy for medium and heavy congestion at 94\% and 95\% correct, respectively. Misclassifications are concentrated between adjacent classes. The most common confusion is medium predicted as heavy at 5\%, and heavy predicted as medium at 5\%. Confusion between light and heavy is negligible. This pattern is consistent with boundary scenarios where gradual speed reductions and tightening headways make medium and heavy clips visually similar even under manual inspection, while light congestion remains more distinct.

\begin{figure*}[!ht]
\centering
\includegraphics[width=0.98\textwidth]{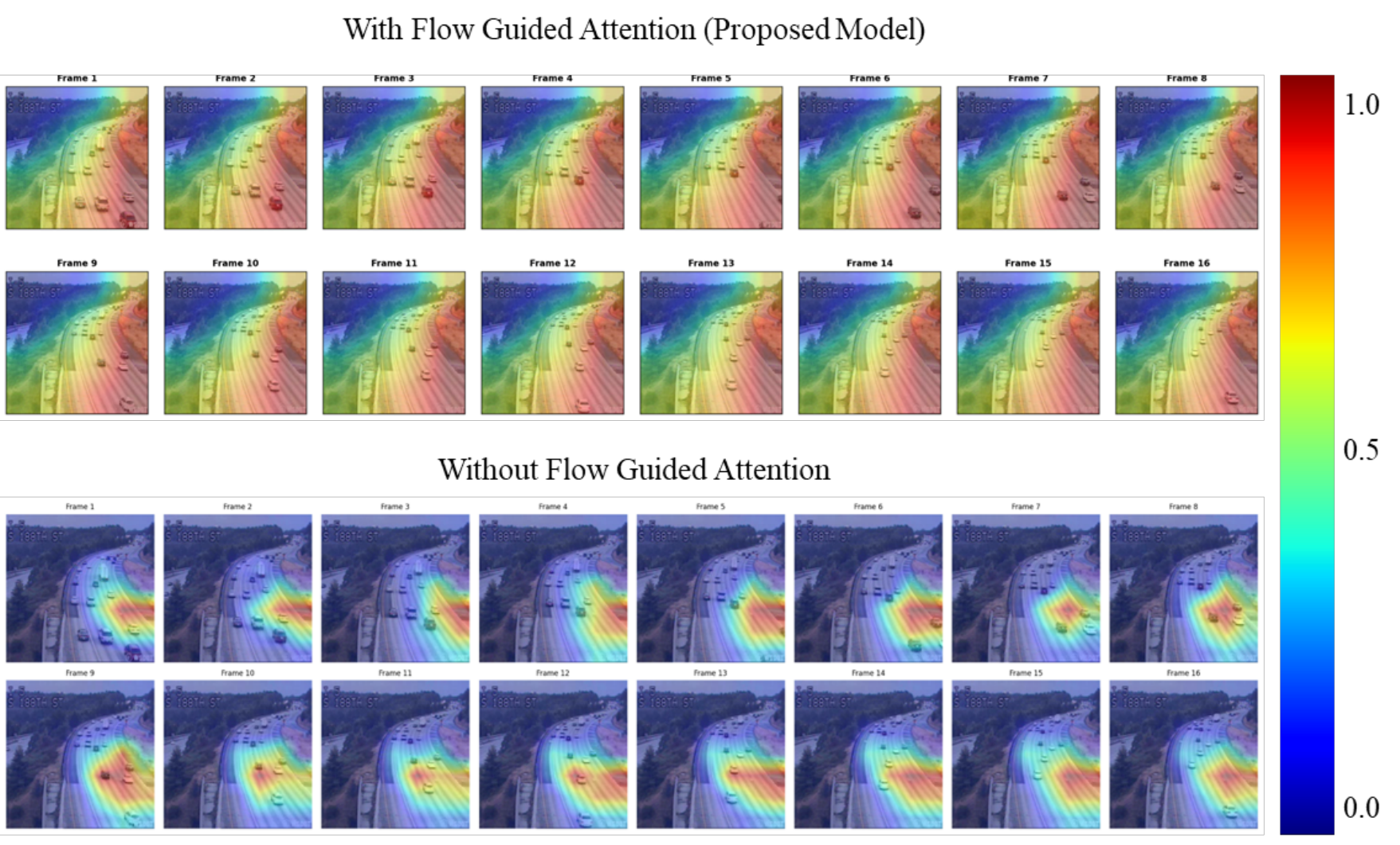}
\caption{Temporal attention visualization analysis across 16 consecutive frames showing attention pattern evolution over time. Top: FLO-EMD with flow-guided attention demonstrating consistent tracking of vehicle locations and adaptive focus on traffic-active regions throughout the sequence. Bottom: FLO-EMD V2 without flow-guided attention exhibiting static attention patterns concentrated on road infrastructure elements with minimal adaptation to dynamic traffic changes. The color scale represents attention weights where warmer colors (red/yellow) indicate higher significance and cooler colors (blue/green) represent lower importance.}
\label{fig:temporal_attention_analysis}
\end{figure*}

Figure~\ref{fig:temporal_attention_analysis} evaluates interpretability by comparing the temporal evolution of attention maps produced by FLO-EMD and the non-guided attention variant. FLO-EMD yields attention that is more temporally coherent and more frequently aligned with traffic-active regions, indicating that motion conditioning reduces the tendency of appearance-driven attention to drift toward visually salient but static infrastructure. However, the visualization also highlights an important limitation: in some sequences, the road-shoulder region is assigned high attention despite little or no vehicle motion in that area. This suggests that the attention mechanism can be influenced by motion-adjacent cues and optical-flow artifacts near boundaries, as well as high-contrast structures such as lane edges, guardrails, shadows, or reflections. As a result, the heatmaps should be interpreted as indicators of model sensitivity rather than direct proxies for vehicle presence, and additional constraints may be required for these visualizations to be used reliably in operational settings.

\section{Conclusion}

This paper presented FLO-EMD, a hybrid congestion-classification framework that integrates motion-based signal analysis with vision-based spatiotemporal representation learning. The method treats dense optical flow as a primary source of congestion dynamics, using it in two complementary roles: first, to condition spatial and channel attention so that appearance features are refined using motion evidence; and second, to form compact optical-flow motion traces whose temporal evolution is decomposed with Empirical Mode Decomposition to capture non-stationary traffic patterns without imposing fixed frequency assumptions. By fusing the learned visual embedding with the EMD-derived motion embedding, FLO-EMD addresses key limitations of prior sensor-driven approaches that lack spatial context and vision-driven approaches whose attention is often appearance-dominated and whose temporal modeling can be insufficiently adaptive.

Experimental results on a multi-source dataset spanning diverse camera viewpoints and environmental conditions demonstrated that FLO-EMD achieves consistent improvements over established 3D CNN, transformer-based, and hybrid baselines under a unified training protocol. Ablation and sensitivity analyses further confirmed that flow-guided attention and EMD provide complementary gains, and that decomposing optical-flow motion traces into intrinsic mode functions improves robustness and generalization relative to variants that rely only on learned temporal encoders. Stratified evaluation across weather and visibility conditions showed stable performance across operational scenarios, with the largest degradation occurring in rain, consistent with the sensitivity of motion cues and appearance features to reflections and reduced contrast.

Despite these results, several limitations warrant acknowledgment. First, the current study primarily focuses on highway traffic scenarios captured from stationary camera viewpoints, which may limit generalizability to urban intersections and complex road geometries. The fixed camera perspective constrains the analysis to specific viewing angles and may not fully represent multi-directional flow patterns that are common in urban networks. Second, although video-based sensing can expand coverage relative to point detectors, camera deployments are also spatially limited for many agencies and are frequently co-located with other sensing technologies, which can reduce practical coverage gains from video alone. In addition, attention maps can occasionally emphasize motion-adjacent regions (for example, shoulders) due to flow artifacts or boundary effects, which can reduce interpretability for operators.

Future work will focus on extending evaluation to intersection-dense urban corridors, improving robustness under precipitation and glare, and introducing attention constraints and calibration strategies that better align visual explanations with traffic-relevant entities. Overall, the results indicate that combining motion-conditioned spatial analysis with adaptive temporal decomposition is an effective direction for reliable congestion classification in traffic surveillance systems.

\section{Data Availability Statement}

The traffic video data used in this study were obtained from four sources: (1) the Interstate 5 (Seattle, Washington) traffic dataset originally compiled by the Statistical Visual Computing Lab (UCSD), which is publicly available at \url{http://visal.cs.cityu.edu.hk/downloads/} \cite{chan2005probabilistic}; (2) the AI-City Challenge 2021 Track 4 dataset, available via \url{https://www.aicitychallenge.org/} \cite{naphade2021aicity}; (3) publicly accessible traffic camera streams from the Tennessee Department of Transportation (TDOT) SmartWay platform (\url{https://smartway.tn.gov/}) \cite{TDOTSmartWay}; and (4) publicly viewable CCTV feeds from the New York State 511NY traffic surveillance network (\url{https://511ny.org/cctv}) \cite{511NY}. Some or all data, models, or code that support the findings of this study are available from the corresponding author upon reasonable request.

\section{Author Contributions}
Eugene Kofi Okrah Denteh contributed to the conceptualization of the research framework, development of the methodology, software implementation, formal analysis of results, data curation, preparation of the original draft manuscript, and creation of visualizations. Blessing Agyei Kyem contributed to the methodology development, software implementation, validation of results, and manuscript review and editing. Joshua Kofi Asamoah contributed to data curation, validation of experimental results, formal analysis, and manuscript review and editing. Armstrong Aboah contributed to the research conceptualization, methodology development, provision of resources, manuscript review and editing, supervision of the research and project administration.

\bibliography{Mybib}

\end{document}